\documentclass[10pt, conference, compsocconf]{IEEEtran}
\usepackage{graphicx}
\usepackage{graphicx}
\usepackage{multirow}
\usepackage{rotating}
\usepackage{algpseudocode}
\usepackage{array}
\usepackage[cmex10]{amsmath}
\begin{document}
\title{Simultaneous Optimization of Neural Network Weights and Active Nodes using Metaheuristics}
\author{
    \IEEEauthorblockN{Varun Kumar Ojha\IEEEauthorrefmark{1}, Ajith Abraham\IEEEauthorrefmark{1},V{\'{a}}clav Sn{\'{a}}{\v{s}}el\IEEEauthorrefmark{1}}
    \IEEEauthorblockA{\IEEEauthorrefmark{1}IT4Innovations, V{\v{S}}B Technical University of Ostrava, Ostrava, Czech Republic
    \\varun.kumar.ojha@vsb.cz,~ajith.abraham@ieee.org,~vaclav.snasel@vsb.cz}
}
\maketitle

\begin{abstract}
Optimization of neural network (NN) significantly influenced by the transfer function used in its active nodes. It has been observed that the homogeneity in the activation nodes does not provide the best solution. Therefore, the customizable transfer functions whose underlying parameters are subjected to optimization were used to provide heterogeneity to NN. For the experimental purpose, a meta-heuristic framework using a combined genotype representation of connection weights and transfer function parameter was used. The performance of adaptive Logistic, Tangent-hyperbolic, Gaussian and Beta functions were analyzed. In present research work, concise comparisons between different transfer function and between the NN optimization algorithms are presented. The comprehensive analysis of the results obtained over the benchmark dataset suggests that the Artificial Bee Colony with adaptive transfer function provides the best results in terms of classification accuracy over the particle swarm optimization and differential evolution.
\end{abstract}


\IEEEpeerreviewmaketitle

\section{Introduction}
\label{sec:intro}
Due to the property of being robust and adaptive with the problem environments, the Neural Network (NN) has emerged as the most desirable computational tool for solving nonlinear and complex optimization, pattern recognition, function approximation classification, etc., problems \cite{haykin1994neural,bishop2006pattern}. On the other hand, the meta-heuristic algorithms are well appreciated for their role in the optimization of the Neural Networks (NNs) \cite{yao1999evolving}. The conventional NN optimizations/training algorithms are efficient in local search or in other words, they are efficient in the exploitation of the current solutions for the creation of new solutions. Whereas, the meta-heuristic algorithms are efficient in both exploitation of the current solution and exploration of the given search space for the creation of new solutions. The meta-heuristic algorithms can be used for the optimization of the connection (synaptic) weights, architecture (geometrical arrangement of the nodes), transfer (activation) functions associated with the nodes and the learning mechanisms \cite{haykin1994neural}.

Yao \cite{yao1999evolving} has summarized the Evolutionary Algorithm (EA) based optimization of the NN where it can be found that the NN optimization is not only limited to optimization of connection weights, but it encompasses the optimization of network architecture, activation function, and learning rules. In the present research, we have illustrated the meta-heuristic framework for the optimization of NN and investigated the impact of the optimization of the underlying parameters of the transfer function associated with the active nodes (nodes at hidden and output layers) of the neural network. In the past, efforts to optimize the transfer functions were mostly limited to finding the appropriate combinations of different variety of transfer functions at the active nodes of a NN. Liu and Yao \cite{Liu1996} have chosen a combination of Sigmoid (Logistic) and Gaussian function to optimize transfer functions of the NN. Similarly, White and Ligomenides \cite{white1993gannet} opted to combine 80\% of Sigmoid and 20\% of Gaussian activation function. Castelli and Trentin \cite{castellipreliminary} have illustrated a connectionist model for an adaptive selection of the transfer function. In their model, they selected hidden unit with a pair  $\{f, p\}$ where $f$ was a set of various transfer functions and $p$ was the corresponding probabilistic measure of the likelihood of the node that was relevant to the computation of the output over the current input. Alimi \cite{alimi2003beta} and \cite{betaDhahri2012designing} have illustrated the significant benefits of using the Beta function in the optimization of the NNs.

In the present research, we have chosen Logistic, Tangent-hyperbolic, Gaussian and Beta function in the optimization of NN. Unlike the research referred above, in the present research, we were inclined toward the homogeneity in NN active nodes. In the other words, all the active nodes in the NN were set using similar transfer functions. Interestingly, in our experiment we have chosen to optimize the parameters of the transfer function that were set at the active nodes of the NN. Thereby, it imparted heterogeneity between the transfer functions of the NN. A similar approach was adopted by van Wyk and Engelbrecht                           \cite{vanWykAndEngelbrecht2011} for the optimization of lambda-gamma NN using particle swarm optimization. We have extended the idea, where we use various transfe function and meta-heuristic algorithms in order to enhance the performance of NN. A meta-heuristic NN optimization framework illustrated in the present paper was used for the simultaneous optimizations of the NN connection weights and the transfer functions. A comprehensive experimental result presented in the paper suggest that setting the active nodes of the NN using the customizable transfer functions whose parameters were optimized using the meta-heuristic algorithms was worth investing efforts. The Artificial Bee Colony (ABC) algorithm used for the simultaneous optimization of the connection weights and the transfer functions was consistent in producing the best result in terms of accuracy in classification of the three classification problem chosen in the present research work.

The rest of the paper is organized as follows: In section \ref{sec:ann}, we discussed the fundamental concept of the NNs, the transfer function and the meta-heuristic algorithms. A discussion on the experimental design and the meta-heuristic framework for the simultaneous optimization of the NN connection weights and the transfer function parameters is provided in section \ref{sec:mhFrame} followed by the results and discussion in section \ref{sec:rnd}. Finally, a conclusion is provided in section \ref{sec:Con}.

\section{Neural Network}
\label{sec:ann}
Artificial Neural Network (NN) or simply the NN imitates the functioning of human brain basically, the biological nervous system that is a network of the immense interconnections between the vast numbers of biological neurons \cite{haykin1994neural}. The neurons are the smallest processing unit of the nervous system. Similarly, the NN is a network of several processing elements (nodes) which gains its capability of behaving intelligently by meticulous training provided using training examples. The NNs have three basic components, architecture, connection weights and learning-rules. In the present scope of the research, we have chosen feed-forward multilayer NN. Geometrically, the NNs are arranged in layer by layer basis, where, each layer may contain one or more computational nodes. Mathematically, the $j^{th}$ node of a NN may be given as:
\begin{equation}
\label{eq:ann}
y_j = \varphi_j \left( \sum_iw_{ji}x_i - b_j \right), 
\end{equation}
where $y_j$ is output of $j^{th}$ node, $w_{ij}$ is connection weight between $i^{th}$ node and $j^{th}$ node, $x_i$ is $i^{th}$ input, $b_j$ is bias at the $j^{th}$ node and $\varphi_j(.)$ is transfer (activation) function at $j^{th}$ node. Since variable $x$ is input (known) and variable $y$ is output (to be computed), we need information of the other remaining variables using a training process that can help to find the optimal values for the variables. Transfer function $\varphi (.)$ is  function input $x$ and variables $\{t_1, t_2, \ldots, t_n\}$ where $t$ is a parameter of transfer function  $\varphi (.)$. It is worth noticing that variable $t$ is usually kept fixed. In subsequent section, we have discussed various kinds of transfer function that may be used at the nodes of a NN.        
\subsection{Transfer Function (TF)}
\label{sub:tf}
Transfer functions at the active nodes of a NN are used to transfer the net scalar input at an active node to a scalar called activation value or the output value at that node. Consult \eqref{eq:ann}, where $y_j$ on the left hand side is the output value of the $j^{th}$ node evaluated using a transfer function $ \varphi(.)$ shown on the right hand side. Basically, the transfer functions limit the net input value at a node to a certain range of values in order to allow a NN to behave in certain ways rather than letting it to behave indiscriminately. Transfer functions may be linear or non-linear. Mostly, the NN is designed to solve non-linearly separable problems. Hence, non-linear transfer functions such as: Logistic (Sigmoid), Tan-hyperbolic, Gaussian, Beta basis function, etc. are be used.

\emph{Logistic Function:} Logistic function \eqref{eq:log1} also known as sigmoid function is a unipolar function. The Logistic function \eqref{eq:log1} has three parameters $x, \lambda $ and $ \theta$, where the variable $x$ indicates net-scalar input value of a node, $\lambda$ indicates steepness and $\theta$ indicates center of the Logistic function. The variables $\lambda$ and $\theta$ control the behavior of Logistic function. The most conventional approach is to keep the variables $\lambda$ and $\theta$ value fixed to one and zero respectively. However, the behavior of the function significantly varied with the variation of the parameters $\lambda$ and $\theta$. Therefore, approaches to optimize the parameters together with the connection weights of a NN will help in obtaining optimal NN. 
\begin{equation}
\label{eq:log1}
\varphi(x,\lambda,\theta) = \frac{1}{1 + e^{(- \lambda (x - \theta))} }
\end{equation}

\emph{Tangent-hyperbolic Function:} Tan-hyperbolic function defined in \eqref{eq:tan} receives parameters $x, \lambda $ and $ \theta$ that are analogues to Logistic functions. However, unlike the unipolar Logistic function, the Tangent-hyperbolic functions are bipolar functions that produce significantly different impact on the net output of a NN with respect to the one produced by the Logistic functions. The parameter steepness $\lambda$ and center $\theta$ controls the behavior of Tangent-hyperbolic function. Hence, optimum values of these parameters may significantly enhance performance of a NN.  
\begin{equation}
\label{eq:tan}
\varphi(x,\lambda,\theta) = \frac{e^{(\lambda (x - \theta))} - e^{(-\lambda (x - \theta))}}{e^{(\lambda (x - \theta))} + e^{(-\lambda (x - \theta))} }
\end{equation}

\emph{Gaussian Function:} Gaussian function \eqref{eq:gauss} parameterized by variables $x, \lambda $ and $ \theta$, where $x$ is net-input and variables $\sigma$ and $\mu$ are width and mean (center) of the function respectively. Gaussian function is a unipolar function that produces a symmetric shape around center $\mu$. The width and center significantly influence the behavior of Gaussian function. Therefore, optimum values of the parameters will be able to enhance the overall performance of a NN.   
\begin{equation}
\label{eq:gauss}
\varphi(x,\sigma,\mu) =  \frac{1}{\sqrt{2 \pi}\sigma}e^{\frac{-(x- \mu)}{2\sigma^{2}}}
\end{equation}

\emph{Basis Function:} Due to Beta function \eqref{eq:betaTf} flexibility and universal approximation characteristics and ability adopt variety different shapes, Alimi \cite{alimi2003beta} used Beta function as an activation function of NN. The Beta function is defined as:
\begin{equation}
\label{eq:BetaBasic}
\beta(x,x_0,x_1,p,q) = \left\lbrace
\begin{matrix}
\left(\frac{x - x_0}{\theta - x_0} \right)^p\left(\frac{x_1 - x}{x_1 - \theta} \right)^q & \mbox{if $x \in ]x_0,x_1[$}\\
0 & \mbox{Otherwise}\\
\end{matrix}
\right.
\end{equation}
where $p>0, q >0 x_0, x_1$ are real parameters $x_0 < x_1$ and $\theta = (p x_1 + q x_0)/(p+q)$ is center of Beta function.
Let  $\sigma = x_1 -x_0$ is the width of the Beta function which can be seen as scale factor for distance like $\parallel x - \theta \parallel $. Hence, $x_0$ and $x_1$ defined as:
\begin{equation}€
\label{eq:BetaParam}
\begin{matrix}
x_0 = \theta - \frac{\sigma p}{p+q} \\ 
x_1 = \theta + \frac{\sigma q}{p+q}
\end{matrix}
\end{equation}
From \eqref{eq:BetaBasic} and \eqref{eq:BetaParam} the Beta function is written as  
\begin{equation}
\label{eq:betaTf}
\varphi(x,\theta,\sigma,p,q) = \left\lbrace
\begin{matrix}
A. B & \mbox{if $x \in ]\theta - \frac{\sigma p}{p+q},\theta + \frac{\sigma q}{p+q}[$}\\
0    & \mbox{Otherwise}\\
\end{matrix}
\right.
\end{equation}
where $A = \left[1 + \frac{(p+q)(x - \theta)}{\sigma p} \right]^p$ and $B = \left[1 - \frac{(p+q)(x - \theta)}{\sigma q} \right]^q $
A detailed discussion on various other types of transfer function provided by Duch and Jankowski \cite{duch1999survey} supports our discussion that an activation functions with various shapes and behaviors influence the overall performance of a NN. Therefore, the necessity of optimizing parameters in the optimization of NN is evident. The conventional NN optimization algorithms uses fixed transfer function. Hence, meta-heuristic optimization algorithms provides a robust platform for the optimization of both the connection weights and the transfer function parameters of a NN.
 
\subsection{Meta-heuristic Algorithms}
\label{sec:Swarm}
Meta-heuristic algorithms are stochastic procedures that are efficient in both exploitation of the present solutions and exploration of the given search space. The meta-heuristic algorithms such as: Artificial Bee Colony, Particle Swarm Optimization and Differential Evolution can be used for optimization of both connection weights and transfer function parameter simultaneously.

\emph{Artificial Bee Colony (ABC):} ABC proposed by Karaboga \cite{abcKaraboga2005} is a meta-heuristic algorithm inspired by foraging behavior of honey bee swarm. The ABC algorithm uses the population of bees to explore the given search space in order to find the optimal solution for a given problem. The ABC algorithm works as follows:  At first a memory of initial food position (candidate solution) is initialized and then the food position is updated by the artificial bees in iterative fashion. The $i^{th}$ candidate in a memory of $P$ solutions, where each of which has $M$ variables can be given as:
\begin{equation}
\label{eq:abcUpdate}
x_{ij} = x_{ij} + rand(-1,1) \times (x_{ij} - x_{kj})
\end{equation}
where, $k \in [1, P]$, and $j \in [i, M]$ and $x_{ij}$ is the comparison between the $i^{th}$food source and a randomly chosen neighbor $k$. A food source is abundant if it is not of good quality and hence a new food source is obtained as:
\begin{equation}
\label{eq:abcNewFood}
x_{i,j} = x_{min,j} + rand(0,1)(x_{max,j} - x_{min,j}),
\end{equation}
where $min$ and $max$ is the bound of the $j^{th}$ variable. Karaboga and Basturk \cite{abcKaraboga2008} have illustrated the application of the ABC algorithm for the optimization of NN connection weights that in our experiment was extended to the optimization of both connection weights and the transfer function simultaneously. Similarly, we used Particle Swarm Optimization and Differential Evolution algorithms for the same purpose.

\emph{Particle Swarm Optimization (PSO):} PSO \cite{psoEberhart1995} is a population based meta-heuristic algorithm imitates the mechanisms of the foraging behavior of swarms. The PSO depends on the velocity and position update of a swarm. The velocity in PSO is updated in order to update the position of the particles in a swarm. Therefore, the whole population moves towards an optimal solution. The PSO uses a population of motile candidate particles characterized by their position $x_i$ and velocity $v_i$ inside the $n-$dimensional search space. Each particle remembers the best position (in terms of fitness function) it visited $b_i$ and knows the best position discovered so far by the whole swarm $g$. At each iteration, the velocity of a particle $i$ is updated according to~\cite{compIntEngelbrecht}:   
\begin{equation}
\label{equ:pso-v}
v_i^{t+1} = c_0 v_i^t + c_1 r_1^t (b_i - x_i^t) + c_2 r_2^r (\bar g^t - x_i^t),
\end{equation}
where $c_1$ and $c_2$ are positive acceleration constants, $r_1$ and $r_2$ are vectors of random values sampled from a uniform distribution, vector $b_i^t$ represents the best position known to particle $i$ at iteration $t$, vector $\bar y^t$ is the best position visited by the swarm at time $t$ and inertia factor $c_0$ is computed as:
\begin{equation}
c_0 = c_0^{max} - \frac{(c_0^{max} - c_0^{min}) \times Current_{iteration}}{Max_{iteration}}.
\end{equation} 
The position of particle $i$ is updated by~\cite{compIntEngelbrecht}:
\begin{equation}
\label{equ:pso-x}
x_i^{t+1} = x_i^t + v_i^{t+1}  
\end{equation}

\emph{Differential Evolution (DE):} DE proposed by \cite{deStorn95} is a popular meta-heuristic algorithm for the optimization of continuous functions. DE has been successfully used for the optimization of NN \cite{deSlowik2011application}. The basic principle of DE is as follows: At first an initial population of $n$ dimensional solutions $x_i$ is constructed. The contraction of new solution takes place iteratively. For the purpose of the construction of new solution, three distinct solutions $a, b$ and $c$ are chosen. Thereafter, a random index $N \in [1,n]$ is chosen. Hence, a new solution $y_i$ is constructed as: 
\begin{equation}
\label{eq:deUpdate}
y_i = \left\lbrace 
\begin{matrix}
a_i + F \times (b_i - c_i) & \mbox{if } r_i < \text{CR or } i = N\\
x_i & \mbox{Otherwise}\\
\end{matrix}
\right.
\end{equation}
where CR indicates the crossover rate, F indicates the weight factor and $r_i$ is a uniform random sample chosen in $(0,1)$.

\section{Meta-heuristic Framework for Transfer function optimization}
\label{sec:mhFrame}
\begin{figure}
\centering
\includegraphics[width=0.4\textwidth]{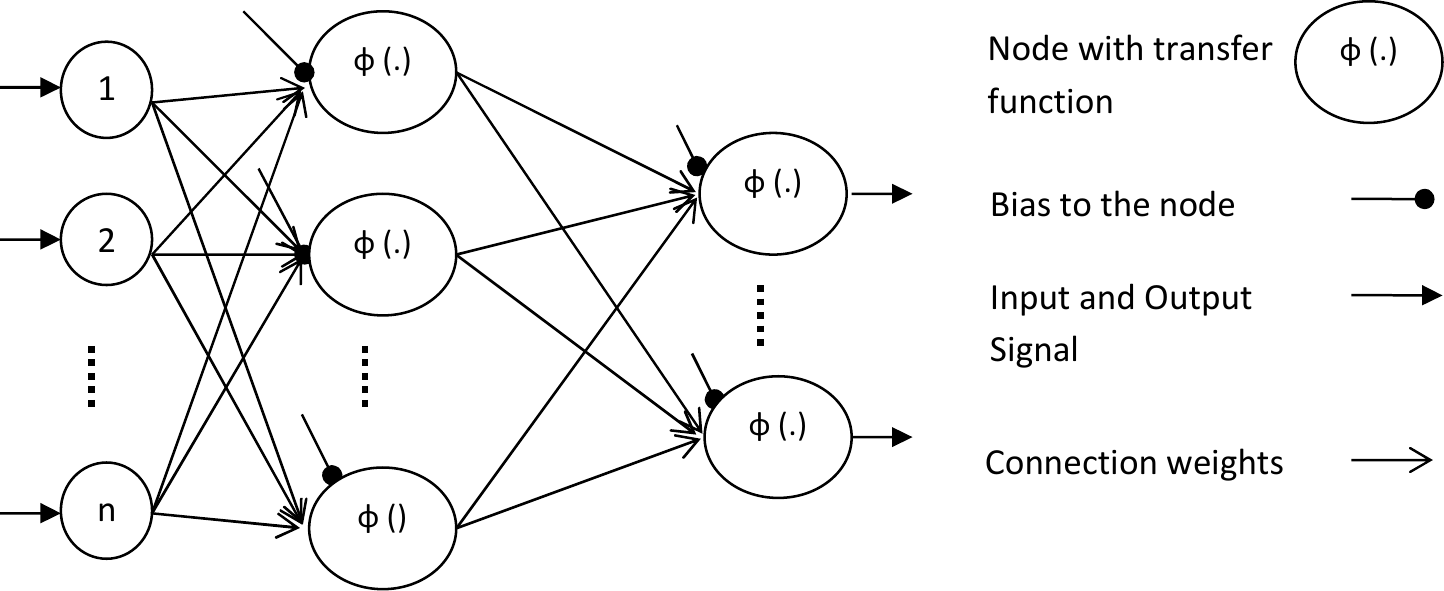}
\caption{Phenotype representation of NN}
\label{fig:nn}
\end{figure}

\begin{figure}
\centering
\includegraphics[width=0.45\textwidth]{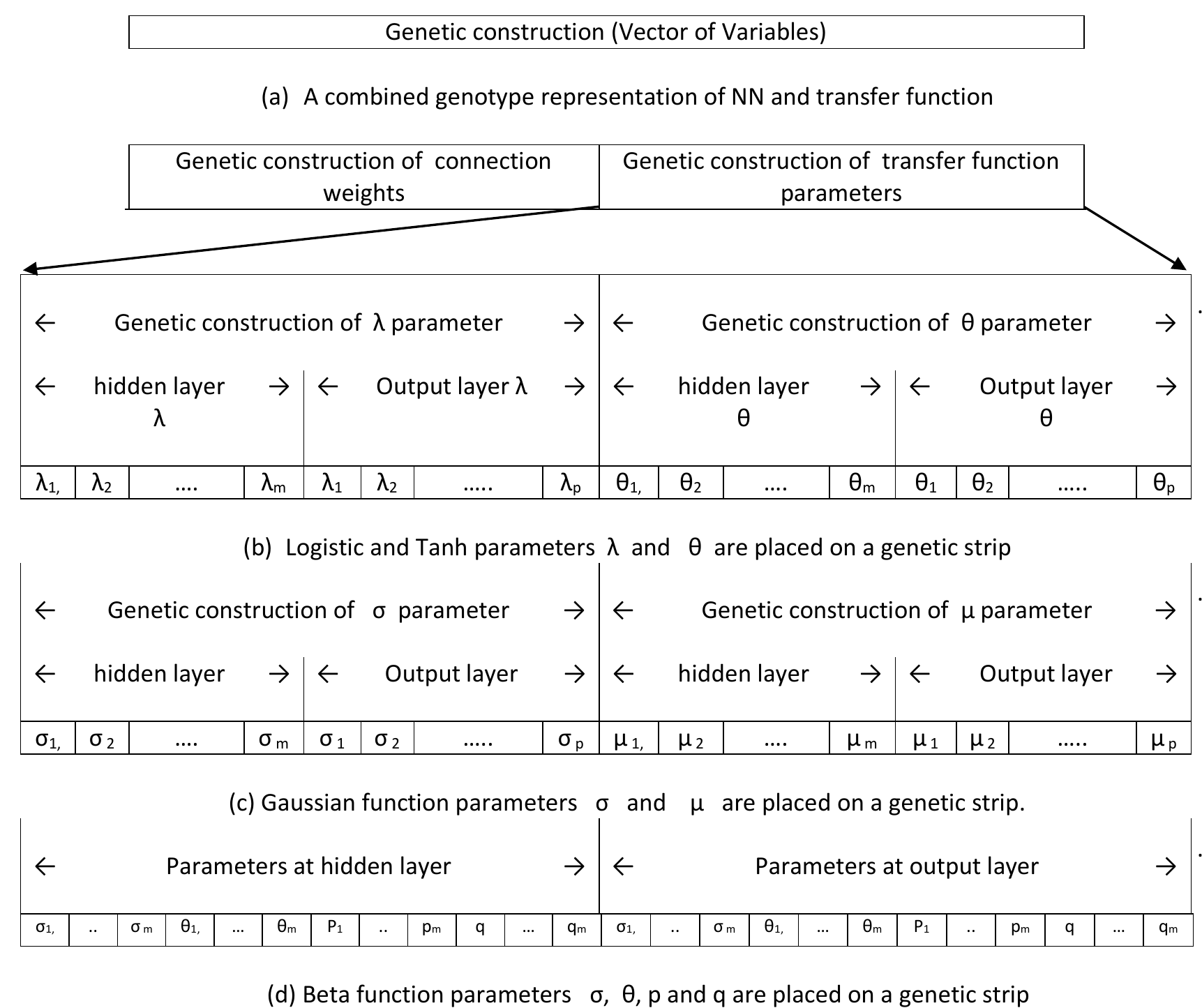}
\caption{Genotype representation of NN}
\label{fig:genotype}
\end{figure}

Meta-heuristic algorithms have proven their competency in optimizing NNs \cite{yao1999evolving,Alba2005} over the conventional NN optimization algorithms such as Backpropagation \cite{annBpRumelhart1986}. In the present research, we have illustrated the role of meta-heuristic algorithms such as: ABC, PSO and DE (version $DE/rand-to-best/1/bin$ \cite{deQin2009}) in the simultaneous optimization of the NN connection weights and transfer function parameters. A three layered feed-forward NN (phenotype) given in Figure \ref{fig:nn} represented as a solution vector (genotype) shown in Figure \ref{fig:genotype} was used for the experiment purpose. It may be noted that, the hidden layer and output layer consist of transfer functions.
Liu and Yao \cite{Liu1996}, Weingaertner \textit{et al.} \cite{weingaertner2002hierarchical} and others \cite{Zhang2010,gomes2013optimization} have adopted heterogeneity in the NN nodes by the means of choosing various transfer function at the hidden layers and the output layers of a NN. On the contrary to their approach, our approach was to explore the impact of the optimization of the individual transfer function parameters on the performance of NN. A meta-heuristic framework for the simultaneous optimization of NN and its transfer function parameters is illustrated in Figure \ref{fig:mhAlgo}, where the meta-heuristic operator $\text{MH}_\text{Operator}$ were defined as per the respective meta-heuristic algorithms. For the experiment, the initial population was constructed using the genotype illustrated in Figure \ref{fig:genotype}. We have chosen the genotype representation with Logistic, Tangent-hyperbolic, Gaussian and Beta basis function. 
\begin{figure}
\begin{algorithmic}[1]
\Procedure{Meta-heuristics-NN}{$\text{W}, \epsilon$} 
\State Initialize  $\text{W}_0$ 
\State Fittest  solution $\text{w}^* = fittest(\text{W}^0)$
 \Repeat
   \State $\text{W}^{t+1} := \text{MH}_\text{Operator}(\text{W}^t)$ 
   \State $\mathrm{\hat{w}}= fittest(\text{W}^{t+1})$ 
   \If{$\mathrm{\bar{w}} < \mathrm{w}^*$}
       \State $\mathrm{w}^* = \mathrm{\bar{w}}$
   \EndIf    
\Until {\textit{ Stopping criteria $\epsilon$ satisfied}}

 \Return $\mathrm{w}^*$
\EndProcedure 
\end{algorithmic}
\caption{Meta-heuristic Framework for Optimization}
\label{fig:mhAlgo}
\end{figure}

The performance of individual meta-heuristic algorithms are subjected to their respective parameter setting. A list of parameter setting used in the experiment for the ABC, PSO and DE is shown in Table \ref{tab:parmSetting}. Apart from the given parameter, the Mersenne-Twister algorithm with random seeds ware used for the initialization of the initial population within a search space $[-1.5,1.5]$.
\begin{table}
\centering
\caption{Parameter Setting of the Algorithms used in the Experiments}
\label{tab:parmSetting}
\begin{tabular}{c | c | c | p{3cm} }
\hline
Algorithm & Population & Iteration & Other\\
\hline
BP  &  10 & 1000 & $\eta = 0.5$ and $m = 0.1$\\
ABC &  10 & 1000 & trial$_{\text{limit}}$ = 100\\
PSO &  10 & 1000 &  $c_1 = c_2 = 2.0$, $c_0^{max} = 1.0$  and $c_0^{min} = 0.0$\\
DE  &  10 & 1000 & CR = 0.9, F = 0.7\\ 
\hline
\end{tabular}
\end{table}      
\section{Results and Discussion}
\label{sec:rnd}
For the experiment purpose, three benchmark dataset (classification problem) from the UCI Machine Learning repository (http://archive.ics.uci.edu/ml/datasets.html) were used. The dataset chosen were Iris, Breast Cancer (Wdbc) and Wine. In order to justify the significance of the proposed model, the primary independent benchmark results given in Table \ref{tab:BpRes} in terms of 10 Cross-Validation (CV) over the mentioned dataset was obtained using BP algorithm \cite{annBpRumelhart1986} that has  learning rate $\eta$ and momentum $m$ as its controlling parameters. The experiment using BP was repeated for the Logistic (SigFix) and Tangent-hyperbolic (TanhFix) function for each of the mentioned dataset. In the experiment using BP algorithm, the parameters $\lambda$ and $\theta$ of the transfer functions SigFix and TanhFix ware set to one and zero respectively.   
\begin{table}
\centering
\caption{The 10 CV results using Backpropagation Algorithm}
\label{tab:BpRes}
\begin{tabular}{l|rr|rr|rr}
\hline
\multirow{2}{*}{Function} & \multicolumn{2}{c|}{Iris} & \multicolumn{2}{c|}{wdbc} & \multicolumn{2}{c}{Wine}\\
\cline{2-7}
         & Error & Var & Error & Var & Error & Var\\
\hline
SigFix  & 0.916 & 0.017 & 0.933 & 0.011 & 0.914 & 0.069\\
TanhFix & 0.956 & 0.009 & 0.945 & 0.012 & 0.955 & 0.017\\
\hline
\end{tabular}
\end{table}

The performance of meta-heuristic algorithms used for the optimization of NN weights and transfer function was tested with the reference to the results obtained using the BP algorithm. Each of the mentioned meta-heuristic algorithms were used optimization of NN independently over the mentioned datasets. The obtained results over Iris, Cancer and Wine classification problems are shown in Tables \ref{tab:resIris}, \ref{tab:resCancer} and \ref{tab:resWine} respectively. The best classification accuracy for the Iris dataset obtained using the BP algorithm with TanhFix was 95.6\%. Whereas, the best classification accuracy using the ABC with Beta function over the same dataset was found to be 98.3\%. It may also be observed from Table \ref{tab:resIris} that optimization of the parameters of the Logistic (SigAdp), Tangent-hyperbolic (TanhAdp), Gaussian and Beta provides classification accuracy better than that of the classification accuracy obtained using the functions with fixed parameter setting. Similarly, the best classification accuracy obtained by the BP over the datasets Cancer and Wine was 94.5\% and 95.5\% respectively and the best classification accuracy over the same dataset using the meta-heuristic algorithms was 97.0\% each. The results obtained over the dataset Iris, Cancer and Wine providing significant evidence that the optimization of functions together with the NN weights helped in obtaining better results than that of using function with fixed parameter setting.  
\begin{table}[!t]
\caption{10CV results on Iris Classification Problem}
\label{tab:resIris}
\centering
\begin{tabular}{l|rr|rr|rr}
\hline
\multirow{2}{*}{Function} & \multicolumn{2}{c|}{ABC} & \multicolumn{2}{c|}{PSO} & \multicolumn{2}{c}{DE}\\
\cline{2-7}
 & \multicolumn{1}{c|}{Error} & \multicolumn{1}{c|}{Var} & \multicolumn{1}{c|}{Error} & \multicolumn{1}{c|}{Var} & \multicolumn{1}{c|}{Error} & \multicolumn{1}{c}{Var}\\
\hline 
SigFix   &         0.774  & 0.248 & 0.799 & 0.271 & 0.729 & 0.190\\
SigAdp   & \textbf{0.972} & 0.014 & 0.839 & 0.669 & 0.859 & 0.322\\
TanhFix  &         0.943  & 0.012 & 0.959 & 0.013 & 0.885 & 0.201\\
TanhAdp  & \textbf{0.978} & 0.008 & 0.759 & 0.150 & 0.846 & 0.179\\
Gaussian & \textbf{0.977} & 0.020 & 0.767 & 0.748 & 0.893 & 0.065\\
Beta     & \textbf{0.983} & 0.008 & 0.839 & 0.188 & 0.944 & 0.074\\
\hline
\end{tabular}
\end{table}
\begin{table}[!t]
\caption{10CV results on Cancer Classification Problem}
\label{tab:resCancer}
\centering
\begin{tabular}{l|rr|rr|rr}
\hline
\multirow{2}{*}{Function} & \multicolumn{2}{c|}{ABC} & \multicolumn{2}{c|}{PSO} & \multicolumn{2}{c}{DE}\\
\cline{2-7}
 & \multicolumn{1}{c|}{Error} & \multicolumn{1}{c|}{Var} & \multicolumn{1}{c|}{Error} & \multicolumn{1}{c|}{Var} & \multicolumn{1}{c|}{Error} & \multicolumn{1}{c}{Var}\\
\hline
SigFix   & \textbf{0.941} & 0.006 &          0.909 & 0.013 &          0.928 & 0.016\\
SigAdp   & \textbf{0.958} & 0.008 & \textbf{0.970} & 0.016 &          0.928 & 0.027\\
TanhFix  & \textbf{0.958} & 0.007 & \textbf{0.957} & 0.011 & \textbf{0.951} & 0.010\\
TanhAdp  & \textbf{0.963} & 0.005 &          0.938 & 0.009 &          0.943 & 0.011\\
Gaussian & \textbf{0.951} & 0.005 &          0.874 & 0.110 &          0.906 & 0.008\\
Beta     & \textbf{0.954} & 0.012 &          0.914 & 0.019 &          0.912 & 0.013\\
\hline
\end{tabular}
\end{table}
Interestingly, the results shown in Tables \ref{tab:resIris}, \ref{tab:resCancer} and \ref{tab:resWine} suggest that the ABC algorithm excels over the other meta-heuristic algorithms such as PSO and DE in the present experiment design with their respective parameter setting mentioned in Table \ref{tab:parmSetting}. However the algorithms PSO and DE have more performance tuning parameters than the ABC. Hence, their performance are subjected to meticulous tuning of their respective parameters. In contrast to van Wyk and Engelbrecht \cite{vanWykAndEngelbrecht2011}, we have performed experiments for the optimized parameters of Tangent-hyperbolic, Gaussian, and Beta function and the results suggests that performance of other adaptive transfer function are better than that of sigmoid function. Similarly, we have used ABC algorithms that performs better than that of PSO algorithm.        
\begin{table}[!t]
\caption{10CV results on Wine Classification Problem}
\label{tab:resWine}
\centering
\begin{tabular}{l|rr|rr|rr}
\hline
\multirow{2}{*}{Function} & \multicolumn{2}{c|}{ABC} & \multicolumn{2}{c|}{PSO} & \multicolumn{2}{c}{DE}\\
\cline{2-7}
 & \multicolumn{1}{c|}{Error} & \multicolumn{1}{c|}{Var} & \multicolumn{1}{c|}{Error} & \multicolumn{1}{c|}{Var} & \multicolumn{1}{c|}{Error} & \multicolumn{1}{c}{Var}\\
\hline
SigFix   & \textbf{0.962} & 0.018 & 0.814 & 0.188 & 0.861 & 0.083\\
SigAdp   & \textbf{0.986} & 0.019 & 0.679 & 0.557 & 0.848 & 0.325\\
TanhFix  & \textbf{0.986} & 0.019 & 0.943 & 0.023 & 0.951 & 0.029\\
TanhAdp  & \textbf{0.990} & 0.015 & 0.873 & 0.031 & 0.895 & 0.061\\
Gaussian & \textbf{0.976} & 0.013 & 0.794 & 0.514 & 0.748 & 0.241\\
Beta     & \textbf{0.970} & 0.028 & 0.867 & 0.058 & 0.871 & 0.046\\
\hline
\end{tabular}
\end{table}

\section{Conclusions}
\label{sec:Con}
In present research, we have presented a meta-heuristic framework for the simultaneous optimization of the NN weights and the parameters of the transfer functions (NN-TFs model). Various types of transfer functions were chosen for the purpose of the rigorous analysis of the influence of the transfer functions optimization together with the NN weights. For the optimization of NN-TFs model, ABC, PSO and DE algorithm were used. Apart from the meta-heuristic algorithms, a Back-propagation algorithm was used for the comprehensive comparison and validation of the significance of the NN-TFs model. The comprehensive results presented in the paper suggests that the adaptive/customizable transfer function helps in enhancing the performance of NN. It may also be observed that the Beta function have four parameters and it is competitive to the Tangent-hyperbolic function that has two controlling parameters. Hence, further examination and tuning of its parameters may offer the best results in comparison to its counterparts. Apart from this, a probabilistic setting for the heterogeneity at the active nodes will be interesting to analyze. From the obtained results, it may also be observed that the ABC excels significantly over the other meta-heuristic such as particle swarm optimization in the present form of their parameter setting.       

\section*{Acknowledgment}
This work was supported by the IPROCOM Marie Curie initial training network, funded through the People Programme (Marie Curie Actions) of the European Union's Seventh Framework Programme FP7/2007-2013/ under REA grant agreement No. 316555.


\bibliographystyle{IEEEtran}
\bibliography{vkoHIS14}
\end{document}